\documentclass{article}
\usepackage{spconf,amsmath,graphicx}
\usepackage{tcolorbox}
\usepackage{tikz}

\usepackage{framed}
\usepackage{algpseudocode}
\usepackage{amsthm,nccmath}
\usepackage{amsmath,bm}
\usepackage[makeroom]{cancel}
\usepackage{bigints}
\usepackage{amssymb}
\usepackage{enumitem}
\usepackage{graphicx}
\usepackage{graphics}
\usepackage{tabularx}
\usepackage{multirow}
\usepackage[font=small,labelfont=bf]{caption}
\usepackage[font=small]{subcaption}
\usepackage{float}
\usepackage{epstopdf}
\usepackage{footnote}
\makesavenoteenv{tabular}
\makesavenoteenv{table}
\usepackage{bbm}
\usepackage{amsthm}
\usepackage{stmaryrd}
\usepackage{amsmath,bm}
\usepackage{amssymb}
\usepackage{mathtools, cuted}
\usepackage{kantlipsum,setspace}
\usepackage[normalem]{ulem}
\usepackage{algorithm,algorithmicx}
\usepackage{amsfonts}
\usepackage{color}
\usepackage{adjustbox}

\usepackage{dblfloatfix}
\theoremstyle{plain}

\newtheorem*{lemma*}{Lemma}

\theoremstyle{definition}

\newtheorem*{defn*}{Definition}

\theoremstyle{remark}

\begin{document}

\title{Anomalous Instance Detection in Deep Learning: A Survey}
\name{Saikiran Bulusu$^1$, Bhavya Kailkhura$^2$, Bo Li$^3$, Pramod K. Varshney$^1$, Dawn Song$^4$}
\address{$^1$Syracuse University, $^2$Lawrence Livermore National Lab, $^3$UIUC, $^4$UC Berkeley\\
\{sabulusu, varshney\}@syr.edu, kailkhura1@llnl.gov, lxbosky@gmail.com, dawnsong@gmail.com
}
\maketitle

\section*{Abstract}
Deep Learning (DL) is vulnerable to out-of-distribution and adversarial examples resulting in incorrect outputs. To make DL more robust, several posthoc anomaly detection techniques to detect (and discard) these anomalous samples have been proposed in the recent past. This survey tries to provide a structured and comprehensive overview of the research on anomaly detection for DL based applications. We provide a taxonomy for existing techniques based on their underlying assumptions and adopted approaches. We discuss various techniques in each of the categories and provide the relative strengths and weaknesses of the approaches. Our goal in this survey is to provide an easier yet better understanding of the techniques belonging to different categories in which research has been done on this topic. Finally, we highlight the unsolved research challenges while applying anomaly detection techniques in DL systems and present some high-impact future research directions.

\section{Introduction}

Deep Learning (DL) techniques provide incredible opportunities to answer some of the most important and difficult questions in a wide range of applications in science and engineering. Therefore, scientists and engineers are increasingly adopting the use of DL for making potentially important decisions in the context of applications of interest, such as {bioinformatics, healthcare, cyber-security, and fully autonomous vehicles}. Several of these applications are often high-regret (i.e., incurring significant costs) in nature.
In such applications, incorrect decisions or predictions have significant costs either in terms of experimental resources when testing drugs, lost opportunities to observe rare phenomena, or in health and safety when certifying parts. 
Most DL methods implicitly assume ideal conditions and rely on the assumption that test data comes from the ``same distribution" as the training data. 
However, this assumption is not satisfied in many real-world applications and virtually all problems require various levels of transformation of the DL output as test data is typically different from the training data either due to noise, adversarial corruptions, or other changes in distribution possibly due to temporal and spatial effects. 
These deviant (or out-of-distribution) data samples are often referred to as anomalies, outliers, novelties in different domains. 
It is well known that DL models are highly sensitive to such anomalies, which often leads to unintended and potentially harmful consequences due to incorrect results generated by DL. Hence, it is critical to determine whether the incoming test data is so different from the training dataset that the output of the model cannot be trusted (referred to as the \textit{anomaly detection} problem). 

Due to its practical importance, anomaly detection has received a lot of attention from statistics, signal processing and machine learning communities. Recently, there has been a surge of interest in devising anomaly detection methods for DL applications. This survey aims to provide a structured overview of recent studies and approaches to anomaly detection in DL based high-regret applications. To the best of our knowledge, there has not been any comprehensive review of anomaly detection approaches in DL systems. Although a number of surveys have appeared for conventional machine learning applications, none of these are specifically for DL applications. This has motivated this survey paper especially in light of recent research results in DL. We expect that this review will facilitate a better understanding of the different directions in which research has been carried out on this topic and potential high-impact future directions.

\subsection{What are Anomalies?}

\begin{figure}[t]
\centering
\includegraphics[width=8cm]{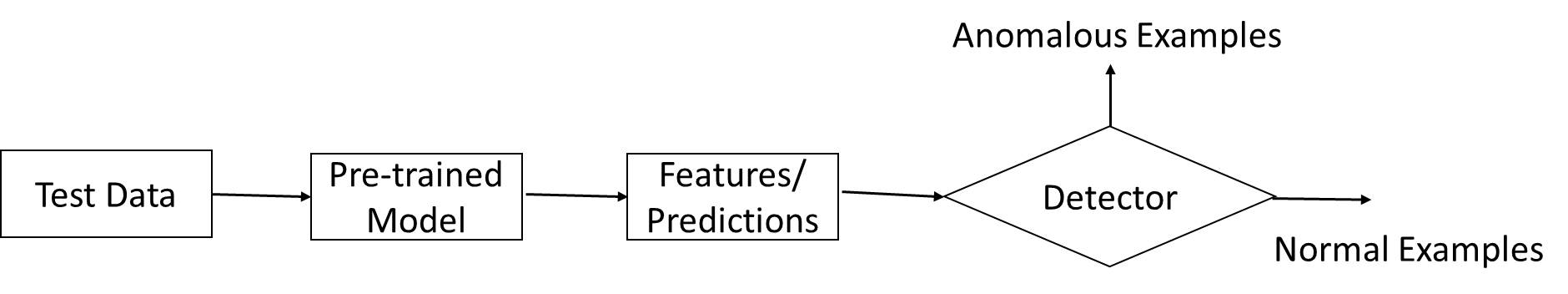}
\caption{Schematic of anomaly detection in DL.}
\label{fig:architecture}
\end{figure}

The problem setup for anomaly detection in deep neural networks (DNNs) is as follows: the DNN is trained on in-distribution data and is asked to perform predictions on both in-distribution as well as out-of-distribution (OOD) test samples. In-distribution test samples are from the same distribution as the training data and the trained DNN is expected to perform reliably on them. On the other hand, anomalous test samples are samples which do not conform to the distribution of the training data. Therefore, predictions of DNNs based on these anomalous samples should not be trusted. The goal of the anomaly detection problem is to design post-hoc detectors to detect these nonconforming test samples (see Fig. \ref{fig:architecture}). 

Next we discuss the types of anomalies, and present their respective differences. We classify anomalies into (a) unintentional and (b) intentional (see Fig. \ref{fig:anomaly}) types.  Unintentional anomalies are independent of the DNN model, as opposed to, intentional anomalies which are intentionally designed by an attacker to force the DNN model to yield incorrect results, and are model dependent.  

\subsubsection{Unintentional: Novel and out-of-distribution examples}
The unintentional anomalies are further classified into novel and OOD examples\footnote{Note that we refer to samples as examples.}. 
Novelty detection is the identification of new or unknown in-distribution data that a machine learning system is not aware of during training. However, the OOD example comes from a distribution other than that of the training data. The distinction between novelties and OOD data is that the novel data samples are typically incorporated into the normal model after being detected, however, OOD samples are usually discarded. In Fig.~\ref{fig:anomaly}, the blue circles outside class boundaries are OOD examples. The OOD examples do not belong to any of the classes. In other words, the classifier is either unaware or does not recognize the OOD examples.

A related problem arises in Domain adaptation (DA) and transfer learning~\cite{zhang2019transfer} which deal with scenarios where a model trained on a source distribution is used in the context of a different (but related) target distribution. The difference between the DA and OOD problems is that DA techniques assume that the test/target distribution is related to the task (or distribution) of interest (thus, utilized during training). On the other hand, OOD techniques are designed to detect if incoming data is so different (and unrelated) from the training data that the model cannot be trusted.

\subsubsection{Intentional: Adversarial Examples}
\begin{figure}[t]
\centering
\includegraphics[width=8cm]{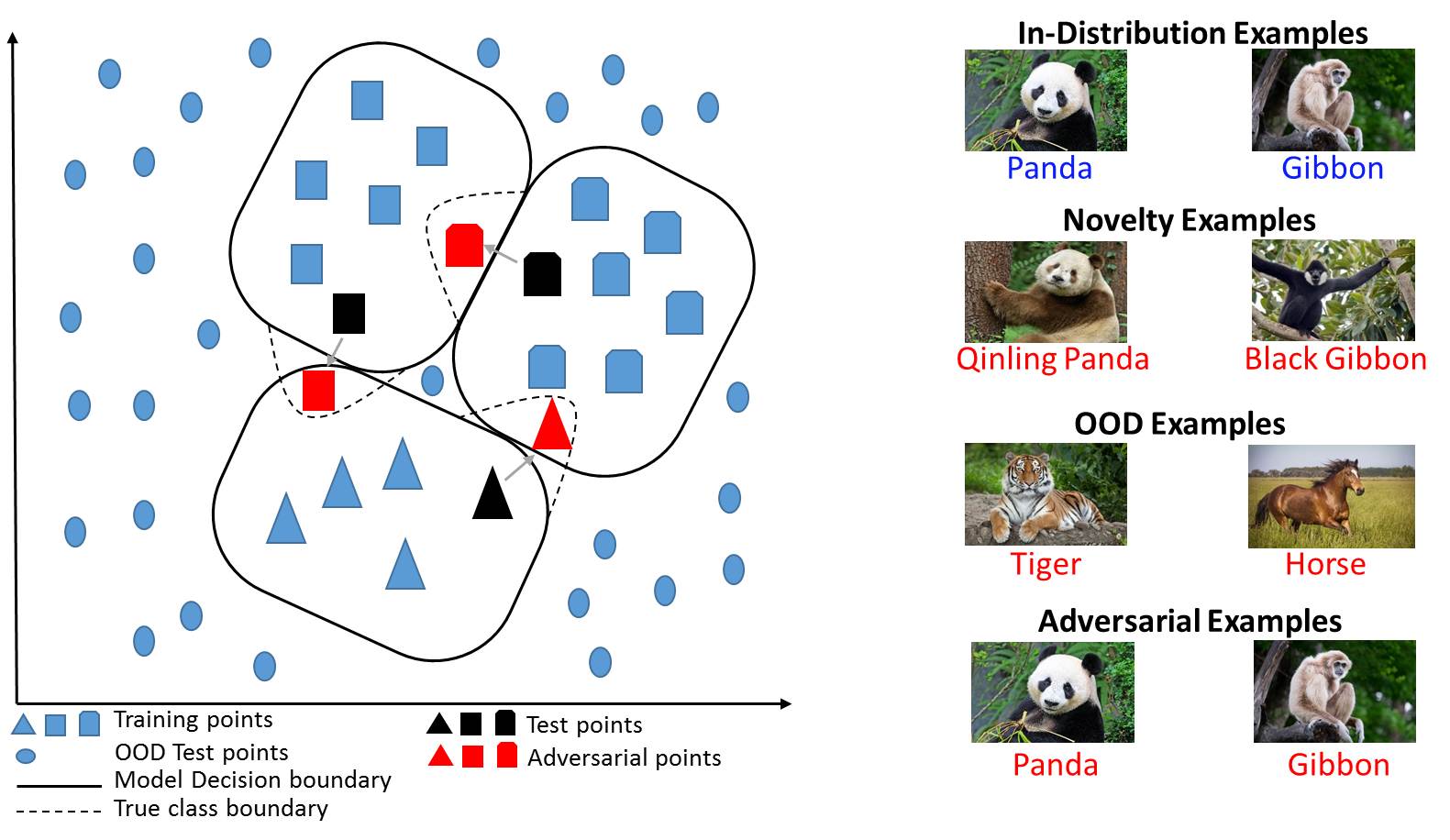}
\caption{ (a) A simple example of anomalies in a 2-dimensional data set. (b) Different variants of anomalous examples for the Panda vs. Gibbon classification problem. Captions indicate the true label and the color indicates whether the prediction is correct or wrong ({blue} for correct and {red} for wrong).}
\label{fig:anomaly}
\end{figure}

The intentional anomalies (also known as the adversarial examples) are the test inputs that are intentionally designed by an attacker to coerce the model to make a mistake. For example, an attacker can modify the input image to fool the DNN classifier which could lead to unforeseen consequences, such as, accidents of autonomous cars or possible bank frauds. In Fig.~\ref{fig:anomaly}, the examples in red are adversarial in nature. Via small perturbation at the input, these examples have been moved to other class regions leading to misclassification. The classifier may or may not have access to some of the labels of these examples leading to different techniques in the literature.  


\subsection{Challenges}
As mentioned above, anomalies are data samples that do not comply with the expected normal behavior. Hence, a naive approach for detecting anomalies is to define a region in the data space that represents normal behavior and declare an example as anomaly if it does not lie in this region. However, there are several factors that make this seemingly simple method ineffective:

\begin{itemize}
    \item The boundary between the normal and anomalous regions is very difficult to define, especially, in complex DNN feature spaces. 
    \item Based on the type of applications, the definition of an anomaly changes. For certain applications, a small deviation in the classification result from that of the normal input data may have far reaching consequences and thus may be declared as anomaly. In other applications, the deviation needs to be large for the input to be declared as an anomaly.
    \item The success of some anomaly detection techniques in the literature depends on the availability of the labels for the training and/or testing data. 
    \item Anomaly detection is particularly difficult when the adversarial examples tend to disguise themselves as normal data. 
\end{itemize}

The aforementioned difficulties make the anomaly detection problem difficult to solve in general. Therefore, most of the techniques in the literature tend to solve a specific instance of the general problem based on the type of application, type of input data and model, availability of labels for the training and/or testing data, and type of anomalies.  

\subsection{Related Work}
Anomaly detection is the subject of various surveys, review articles, and books. In~\cite{Chandola_acm_2009}, a comprehensive survey of various categories of anomaly detection techniques for conventional machine learning as well as statistical models is presented. For each category of detection, various techniques and their respective assumptions along with the advantages and disadvantages are discussed. The computational complexity of each technique is also mentioned. A comprehensive survey of the novelty detection techniques is presented in~\cite{Pimentel_sp_2014}. Various techniques are classified based on the statistical models used and the complexity of methods. Recently, an elaborate survey is presented in~\cite{Chalapathy_corr_2019} where DL based anomaly detection techniques are discussed. Here, two more categories of anomaly detection, namely, hybrid models as well as  one-class DNN techniques are also included. Note that our survey paper is different from \cite{Chalapathy_corr_2019} as our focus is on discussing unintentional and intentional anomalies specifically in the context of DNNs whereas \cite{Chalapathy_corr_2019} discusses approaches which use DNN based detectors applied to conventional ML problems. In some sense, our survey paper is much broader in the context of DL applications. In~\cite{agrawal2015survey}, a survey of the data mining techniques used for anomaly detection are discussed. The techniques discussed are clustering, regression, and rule learning. Furthermore, in~\cite{Salehi2018survey}, the authors discuss the models that are adaptive to account for the data coming from the dynamically changing characteristics of the environment and detect anomalies from the evolving data. Here, the techniques account for the change in the underlying data distribution and the corresponding unsupervised techniques are reviewed. In~\cite{kalinichenko2014methods}, the anomaly detection techniques are classified based on the type of data namely, metric data, evolving data, and multi-structured data. The metric data anomaly detection techniques consider the use of metrics like distance, correlation, and distribution. The evolving data include discrete sequences and time series. In~\cite{patcha2007overview}, various statistical techniques, data mining based techniques, and machine learning based techniques for anomaly detection are discussed. In~\cite{hodge2004survey,Muruti2018survey}, the existing techniques for anomaly detection which include statistical, neural network based, and other machine learning based techniques are discussed. Various books~\cite{dunning2014practical,mehrotra2017anomaly,aggarwal2016outlier,bhuyan2017network} also discussed the techniques for anomaly detection. 

\subsection{Our Contributions}

To the best of our knowledge, this survey is the first attempt to provide a structured and a broad overview of extensive research on detection techniques spanning both unintentional and intentional anomalies in the context of DNNs. 
Most of the existing surveys on anomaly detection focus on (i) anomaly detection techniques for conventional machine learning algorithms and statistical models, (ii) novelty detection techniques for statistical models, (iii) DL based anomaly detection techniques.
In contrast, we provide a focused survey on post-hoc anomaly detection techniques for DL.
We classify these techniques based on the availability of labels for the training data corresponding to anomalies, namely, supervised, semi-supervised, and unsupervised techniques. We discuss various techniques in each of the categories and provide the relative strengths and weaknesses of the approaches. We also briefly discuss anomaly detection  techniques that do not fall in the post-hoc category, e.g.,training-based, architecture design, etc.   

\begin{figure}[t]
\centering
\includegraphics[width=8cm]{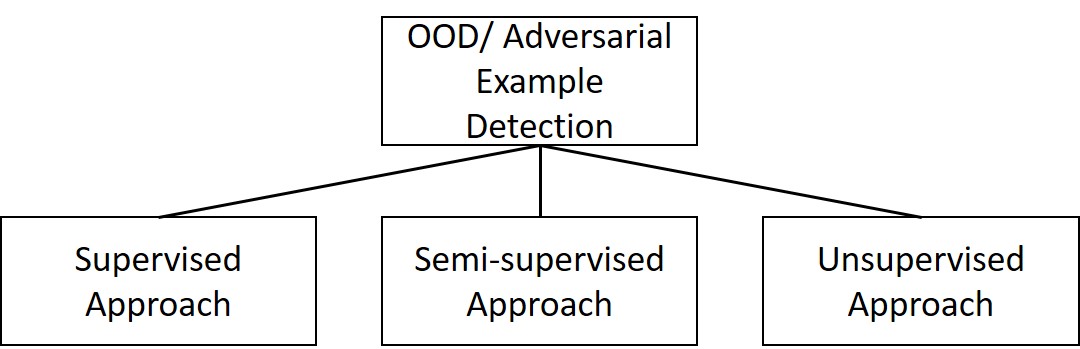}
\caption{Schematic representation of different types of anomaly detection techniques discussed in this survey.}
\label{fig:outline}
\end{figure}

\subsection{Organization}
This survey is organized mainly in three parts: detection of unintentional anomalies, detection of intentional anomalies, and applications. For both unintentional and intentional anomalies, we will discuss different types of approaches (as illustrated in Fig.~\ref{fig:outline}).
In Sec. 2, we present various post-hoc anomaly detection techniques which are used to detect unintentional anomalies. These techniques are classified based on the availability of labels. 
In Sec. 3, we present various post-hoc anomaly detection techniques which are used to detect intentional anomalies (or adversarial examples). The techniques are again classified based on the availability of labels. 
In Sec. 4, we discuss strengths and weaknesses of different categories of methods.
In Sec. 5, we describe various application domains where anomaly detection is applied. 
Finally, we conclude and present open questions in this area in Sec. 6.

\section{Unintentional Anomaly Detection}

In this section, we discuss the detection techniques which detect the OOD examples given a pre-trained neural network. Most DL approaches assume that the test examples belong to the same distribution as the training examples. Consequently, the neural networks are vulnerable to test examples which are OOD. Hence, we need techniques to improve the reliability of the predictions or determine whether the test example is different in distribution from that of the training dataset. Here, we concentrate on the techniques that determine whether the test example is different in distribution from that of the training dataset, using the pre-trained DNN followed by a detector. We refer to this architecture as post-hoc anomaly detection. A topic related to OOD example detection is novelty detection~\cite{domingues2019comparative,marsland2003novelty,bouguelia2018adaptive} which aims at detecting previously unobserved (emergent, novel) patterns in the data. It should be noted that solutions for novelty detection related problems are often used for OOD detection and vice-versa, and hence we use these terms interchangeably in this survey.
Based on the availability of labels for OOD data, techniques are classified as supervised, semi-supervised, and unsupervised which are discussed next and summarized in Table~\ref{tab:oodtable}.

\begin{table}[!t]
\caption{OOD detection related papers.}
\begin{adjustbox}{width=\columnwidth,center}
 \begin{tabular}{||c c c||} 
 \hline
 Classification Type & Reference & Contributions  \\ [0.5ex] 
  \hline
 Supervised & \cite{Oberdiek_corr_2018}& \begin{tabular}{@{}c c@{}}Uncertainty measure based on the\\ gradient of the negative log-likelihood\\ is used as a measure of confidence\end{tabular}\\
     \hline
 Supervised & \cite{Lee_nips_2018}& \begin{tabular}{@{}c c@{}} Confidence scores based on Mahalanobis\\ distance from different layers is \\combined using weighted averaging \end{tabular}  \\
   \hline
  Supervised & \cite{Bahat2018ConfidenceFI} & \begin{tabular}{@{}c c@{}} Invariance of classifier's softmax under\\ various transformations to input image\\ is  used as a measure of confidence\end{tabular}\\
    \hline
  Supervised & \cite{Jiang_nips_2018}& \begin{tabular}{@{}c c@{}} Ratio of Hausdorff distances between test\\  sample to the nearest non-predicted and\\  the predicted classes is used as the trust score \end{tabular}   \\
    \hline
   Semi-supervised & \cite{Liu_corr_2018} & \begin{tabular}{@{}c c@{}} Probably Approximately Correct (PAC)\\ algorithm is proposed to guarantee a\\ user-specified anomaly detection rate \end{tabular}  \\
   \hline
 Semi-supervised & \cite{Ren2019LikelihoodRF}  & \begin{tabular}{@{}c c@{}}Likelihood ratio-based method is \\used to differentiate between\\ in-distribution and OOD examples\end{tabular}\\
 \hline
 Semi-supervised & \cite{yu2019unsupervised} & \begin{tabular}{@{}c c@{}} A two-head CNN consisting of a\\ common feature extractor and two \\classifiers with different decision boundaries\\ is trained to detect OOD examples\end{tabular}\\
 \hline
  Unsupervised & \cite{Hendrycks_corr_2016} &  \begin{tabular}{@{}c c@{}}Predicted softmax probability is\\ used to detect OOD examples \end{tabular} \\
  \hline
  Unsupervised & \cite{Liang_icml_2017} & \begin{tabular}{@{}c c@{}}Temperature scaling and by adding small \\perturbations to the input is used to better\\ separate the softmax score for OOD detection \end{tabular}  \\
   \hline
 Unsupervised & \cite{Lawson2017finding}& \begin{tabular}{@{}c c@{}} GAN based architecture is used  to \\compare the bottleneck features of the \\generated image with that of the test image\end{tabular}\\
 \hline
 Unsupervised & \cite{Chen_corr_2018}& \begin{tabular}{@{}c c@{}} Degenerated prior network with\\ concentration perturbation algorithm\\  is used to get better uncertainty measure \end{tabular}  \\
 \hline
 Unsupervised &\cite{golan2018deep}&  \begin{tabular}{@{}c c@{}}Learning to discriminate between geometric \\transformations is used for learning unique \\features  that  are  useful in OOD detection\end{tabular}\\
 \hline
 Unsupervised & \cite{Denouden2018ImprovingRA}& \begin{tabular}{@{}c c@{}} Mahalanobis distance is applied in the latent \\space of the autoencoder to detect OOD examples \end{tabular}\\
 \hline
   Unsupervised & \cite{schulam_jmlr_2019}& \begin{tabular}{@{}c c@{}}Resampling uncertainty estimation approach\\  is proposed as an approximation to the bootstrap \end{tabular} \\
 \hline
 \label{tab:oodtable}
 \end{tabular}
 \end{adjustbox}
\end{table}

\subsection{Supervised Approaches}
\label{supervised}

In this section, we review the anomaly detection approaches when the labels of both the in-distribution and the OOD examples are available to enable differentiation between them as the supervised anomaly detection problem. Any unseen test data sample is compared with the detector to determine which class (in-distribution vs. OOD) it belongs to.

In~\cite{Oberdiek_corr_2018}, an approach to measure uncertainty of a neural network based on gradient information of the negative log-likelihood at the predicted class label is presented. The gradient metrics are computed from all the layers in this method and scalarized using norm or min/max operations. A large value of the gradient metrics indicates incorrect classification or OOD example.  A convolutional neural network (CNN) is used as the classifier trained on Extended MNIST digits~\cite{cohen2017emnist}. EMNIST letters, CIFAR10~\cite{krizhevsky2009learning} images as well as different types of noise are used as OOD data. The authors found that such an unsupervised scheme does not work well on all types of OOD data. Therefore, a supervised variant of this scheme where one allows an anomaly detector to be trained on uncertainty metrics of some OOD samples is proposed. It was shown that the performance is improved considerably by utilizing the labeled OOD data.  

In~\cite{Lee_nips_2018}, the high-level idea is to measure the probability density of test sample on DNN feature spaces. Specifically, the authors fit class-conditional Gaussian distributions to pre-trained features. This is possible since the posterior distribution can be shown to be equivalent to the softmax classifier under Gaussian discriminant analysis. Next, a confidence score using the Mahalanobis distance with respect to the closest class conditional distribution is defined. Its parameters are chosen to be empirical class means and tied empirical covariance of training samples. To further improve the performance, confidence scores from different layers of DNN is combined using weighted averaging. Weight of each layer is learned by training a logistic regression detector using labeled validation samples comprising of both in-distribution and OOD data. The method is shown to be robust to OOD examples. 

In~\cite{Bahat2018ConfidenceFI}, a detector was trained on representations derived from a set of classifier responses generated from applying different natural transformation to a given image. Analyzing the invariance of classifier's decision under various transformations establishes a measure of confidence in its decision. In other words, the softmax values of the OOD input should fluctuate across transformed versions, while those of the in-distribution image should be relatively stable. The authors trained a binary OOD detector on confidence scores under various transformations for in-distribution vs. OOD training data. ResNet based architecture is used as the classifier and the Self-Taught Learning (STL-10) dataset~\cite{coates2011analysis} is used as the in-distribution data and the Street View House Numbers (SVHN) dataset~\cite{netzer2011reading} is used as the OOD data. The approach is shown to outperform other baselines.

In~\cite{Jiang_nips_2018}, a trust score is proposed to know whether the prediction of a test example by a classifier can be trusted. This score is defined as the ratio of the Hausdorff distances between the distance from the testing sample to the nearest class different from the predicted class (e.g., OOD class) and the distance to the predicted class. To compute the trust score, the training data is pre-processed to find a high density set of each class to filter outliers. The trust score is estimated based on this high density set. 
The idea behind the approach is that if the classifier predicts a label that is considerably farther than the closest label, then it may be an OOD or unreliable example. For the task of identifying correctly/incorrectly classified examples, it was shown that the trust score performs well in low to medium dimensions. However, it performs similar to classifiers' own reported confidence (i.e., probabilities from the softmax layer) in high dimensions.


\subsection{Semi-supervised Approaches}
We refer to the anomaly detection techniques as semi-supervised if they utilize unlabeled contaminated data (or information) in addition to labeled instances of in-distribution class. Since, these techniques do not require to know whether unlabeled instance is in-distribution or OOD, they are more widely applicable than supervised techniques.  

In~\cite{Liu_corr_2018}, the algorithm uses the knowledge of the upper bound on the number of anomaly examples in the training dataset to provide Probably Approximately Correct (PAC) guarantees for achieving a desired anomaly detection rate. The algorithm uses cumulative distribution functions (CDFs) over anomaly scores for the clean and contaminated training datasets to derive an anomaly threshold. An anomaly detector assigns score for all the test examples and orders them according to how anomalous the examples are with respect to the in-distribution data. This ordered score vector is then compared to the threshold to detect the OOD examples. The threshold is computed such that it guarantees a specific anomaly detection rate. Empirical results on synthetic and standard datasets show that the algorithm achieves guaranteed performance on OOD detection task given enough data.

In~\cite{Ren2019LikelihoodRF}, a likelihood ratio-based method using deep generative models is presented to differentiate between in-distribution and OOD examples. The authors assumed that the in-distribution data is comprised of both semantic and background parts. The authors found that the likelihood can be confounded by the background (e.g. OOD input with the same background but different semantic component). Using this information about OOD data, they propose to use a background model to correct for the background statistics and enhance the in-distribution specific features for OOD detection. Specifically, background model is trained by adding the right amount of perturbations to inputs to corrupt the semantic structure in the data. Hence, the model trained on perturbed inputs captures only the population level background statistics. This likelihood ratio is computed from the in-distribution data and the background statistics. If the likelihood ratio is larger than a pre-specified threshold, it is highly likely that the test example is OOD. The National Center for Biotechnology Information microbial genome dataset is utilized in~\cite{Ren2019LikelihoodRF} in the following manner. Various bacteria are grouped into classes which were discovered over the years. Specifically, the classes discovered before a given cutoff year are considered as in-distribution classes and those discovered after the cutoff year are considered OOD classes. The proposed test improves the accuracy of OOD detection compared to the accuracy of the state-of-the-art detection results.

In \cite{yu2019unsupervised}, a semi-supervised OOD detection technique based on two-head CNN was proposed. The idea is to train a two-head CNN consisting of one common feature extractor and two classifiers which have different decision boundaries but can classify in-distribution samples correctly. Further, unlabeled contaminated data is used to maximize the discrepancy between two classifiers to push OOD samples outside in-distribution manifold. This enables the detection of OOD samples that are far from the support of the in-distribution samples.



\subsection{Unsupervised Approaches}

We refer to the detection techniques as unsupervised if they only utilize in-distribution data for OOD detection. 

In~\cite{Hendrycks_corr_2016}, as the statistics derived from the softmax distributions are helpful, a baseline method based on softmax to determine whether or not a test example is OOD is proposed. The idea is that a well trained network tends to assign higher predicted probability to in-distribution examples than to OOD examples. Hence, the OOD example can be detected by comparing the predicted softmax class probabilities of the examples to a threshold. Specifically, the authors generated the training data by separating correctly and incorrectly classified test set examples and, for each example, computing the softmax probability of the predicted class which was used to compute the threshold. The performance of this approach was evaluated on computer vision, natural language processing and speech recognition tasks. The technique fails if the classifier does not separate the maximum values of the predictive distribution well enough with respect to in-distribution and OOD examples. 
Therefore, the authors in~\cite{Liang_icml_2017} proposed a method based on the observation that using temperature scaling and adding small perturbations to the input can better separate the softmax score distributions between in- and out-of-distribution images. Wide ResNet~\cite{zagoruyko2016wide} and DenseNet~\cite{huang2017densely} architectures were used and trained using the CIFAR-10 and CIFAR-100~\cite{krizhevsky2009learning} as in-distribution datasets. 
The OOD detector was tested on several different natural image datasets and synthetic noise datasets. It was shown that the approach significantly improves the detection performance and outperforms the baseline in~\cite{Hendrycks_corr_2016}.

In~\cite{Lawson2017finding}, a generative adversarial network (GAN)~\cite{goodfellow2014generative} based architecture is used in reconstruction error based OOD detection method. The motivation is that the GAN will perform better when generating images from previously seen objects (i.e., in-distribution data) than it will when generating images of objects it has never seen before (i.e., OOD data). In this approach, the test image is first passed through the generator of the GAN, which produces bottleneck features and a reconstructed image. Next, the reconstructed image is passed through the encoder producing another set of bottleneck features. The Euclidean distance between these two feature sets represents a measure of how much the generated image deviates from the original image and is used as an anomaly score.

In~\cite{Chen_corr_2018}, the authors propose a degenerated prior network architecture, which can efficiently separate model-level uncertainty from data-level uncertainty via prior entropy. To better separate in-distribution and OOD images, they propose a concentration perturbation algorithm, which adaptively adds noise to concentration parameters of prior network. Through comprehensive experiments, it was shown that this method achieves state-of-the-art performance especially on the large-scale dataset. However, this method is found to be sensitive to different neural network architectures, which could sometimes lead to inferior performance.

In~\cite{golan2018deep}, the intuition is that learning to discriminate between geometric transformations applied to images help in learning of unique features of each class that are useful in anomaly detection. The authors train a multi-class classifier over a self-labeled dataset created by applying various geometric transformations to in-distribution images. At test time, transformed images are passed through this classifier, and an anomaly score derived from the distribution of softmax values of the in-distribution training images is used for detecting OOD data. The classifier used is the Wide Residual Network model~\cite{zagoruyko2016wide} trained on CIFAR dataset. The CatsvsDogs dataset~\cite{elson2007asirra}, that contains 12,500 images of cats and dogs each, is treated as the OOD data. The method performs better compared to the baseline approaches in~\cite{Hendrycks_corr_2016} for the larger-sized images and is robust to the OOD examples. The method is able to distinguish between the normal and OOD examples with a significant margin compared to the baseline methods.     

The approach in \cite{Denouden2018ImprovingRA} (and references therein) consider the problem of detecting OOD samples based on the reconstruction error. These methods assume that OOD data is composed of different factors than in-distribution data. Therefore, it is difficult to compress and reconstruct OOD data based on a reconstruction scheme optimized for in-distribution data. Specifically, \cite{Denouden2018ImprovingRA} proposes to incorporate the Mahalanobis distance in latent space to better capture these OOD samples. They combined the Mahalanobis distance between the encoded test sample and the mean vector of the encoded training set with the reconstruction loss of the test sample to construct an anomaly score. 
Single digit class from MNIST~\cite{lecun-mnisthandwrittendigit-2010} is used as in-distribution and the other classes of MNIST are treated as OOD samples. The authors illustrate that by including the latent distance helps in improving the detection of in-distribution and OOD examples.


In~\cite{schulam_jmlr_2019}, the predictions of a pre-trained DNN are audited to determine their reliability. Resampling uncertainty estimation (RUE) approach is proposed as an approximation to the bootstrap procedure. Intuitively, RUE estimates the amount that a prediction would change if different training data was used from the same distribution. It quantifies uncertainty using the gradients and Hessian of the model's loss on training data and bootstrap samples to produce an ensemble of predictions for a test input. This uncertainty score is compared to a threshold for detecting correct and incorrect predictions. A single hidden layer feedforward neural network architecture is trained using eight common benchmark regression datasets~\cite{hernandez2015probabilistic} from the UCI dataset repository. The authors show that the uncertainty score detects inaccurate predictions for auditing reliability compared to existing techniques more effectively. This approach can also be used to detect OOD samples.  

Note that the unsupervised methods discussed above require comparing proposed anomaly scores with a threshold. Although thresholds are computed solely based on in-distribution data, one can further improve the performance by optimally choosing thresholds based on OOD validation samples (if available).

\subsection{Other Miscellaneous Techniques}

In this section, we discuss various approaches that are different from the post-hoc anomaly detection techniques, e.g., training-based, architecture design, etc.

In~\cite{Shilton2013multiclass}, a new form of support vector machine (SVM) is presented that combines multi-class classification and OOD detection into a single step. Specifically, the authors augmented original SVM with an auxiliary zeroth class as the anomaly class for labeling OOD examples. The UCI datasets are used as the training examples. The authors demonstrate the trade-off between the ability to detect anomalies and the incorrect labeling of normal examples as anomalies. 

A hybrid model for fake news detection in~\cite{ruchansky2017csi} consists of three steps which capture the temporal pattern of user activity on a given article using a recurrent neural network (RNN), checking the credibility of the media source, and classifying the article as fake or not. 
In~\cite{filonov2017rnn}, an RNN network is used to detect anomalous data where the Numenta Anomaly Benchmark metric is used for early detection of anomalies. 

The method presented in~\cite{Techapanurak_corr_2019} proposed to modify the output layer of DNNs. Specifically, instead of using logit scores for computing class probabilities, the cosine of the angle between the weights of a class and the features of the class are used. In other words, the class probabilities are obtained using the softmax of scaled cosine similarity. The detection of OOD samples is done by comparing the maximum of cosine values across classes to a threshold. The method is hyperparameter-free and has high OOD detection performance. However, the trade-off is the degradation of the classification accuracy. The Wide Residual Network~\cite{zagoruyko2016wide} is used as the classifier trained using the CIFAR dataset, and tiny ImageNet and SVHN datasets are considered OOD data. The approach achieves competitive detection performance even without the tuning of the hyperparameters and the method requires only a single forward pass without the need for backpropagation for each input.




In~\cite{wang2019effective}, a deep autoencoder is combined with CNN to perform supervised OOD detection. Autoencoder is used as a pre-training method for supervised CNN training. The idea is to reconstruct high-dimensional features using the deep autoencoder and detect anomalies using CNNs. It was shown that this combination can improve the accuracy and efficiency of large-scale Android malware detection.




A novel training method is presented in~\cite{lee2018training} where two additional terms are added in the cross entropy loss that minimize the Kullback-Leibler (KL) distance between the predictive distribution on OOD examples and the uniform distribution to assign less confident predictions to the OOD examples. Then, in-distribution and OOD samples are expected to be more separable. However, the loss function for optimization requires OOD examples for training which are generated by using a GAN architecture. Hence, the training involves minimizing the classifier’s loss and the GAN loss alternately. 

In~\cite{Vyas_corr_2018}, the algorithm comprises of an ensemble of leave-out-classifiers. Each classifier is trained using in-distribution examples as well as OOD examples. Here, the OOD examples are obtained by designating a random subset from the training dataset as OOD and the rest are in-distribution. A novel margin-based loss function is presented that  maintains a margin $m$ between the average entropy of the OOD and in-distribution samples. Hence, the loss function is the cross-entropy loss along with the margin-based loss. The loss function is minimized to train the ensemble of classifiers. The OOD detection score is obtained by combining the softmax prediction score and the entropy with temperature scaling. The score is shown to be high for in-distribution examples and low for OOD examples. 

Furthermore, \cite{Hendrycks_corr_2018} proposes leveraging alternative data sources to improve OOD detection by training anomaly detectors against an auxiliary dataset of outliers, an approach they call Outlier Exposure. The motivation is that while it is difficult to model every variant of anomaly distribution, one can learn effective heuristics for detecting OOD samples by exposing the model to diverse OOD datasets. Thus, learning a more conservative concept of the in-distribution and enabling anomaly detectors to generalize and detect unseen anomalies.

The key idea in~\cite{choi2019generative} is that the likelihood models assign higher density values to the OOD examples than the in-distribution examples. The authors propose generative ensembles to detect OOD examples by combining a density evaluation model with predictive uncertainty estimation on the density model via ensemble variance. Specifically, they use uncertainty estimation on randomly sampled GAN discriminators to de-correlate the OOD classification errors made by a single discriminator.

The authors in \cite{song2019unsupervised} proposed a permutation test statistics to detect OOD samples using deep generative models trained with batch normalization. They show that the training objective of generative models with batch normalization can be interpreted as maximum pseudo-likelihood over a different joint distribution. Over this joint distribution, the estimated likelihood of a batch of OOD samples is shown to be much lower than that of in-distribution samples.

In~\cite{Ovadia_corr_2019}, benchmarking of some of the existing posthoc calibration based OOD detection techniques is performed. The effect of OOD examples on the accuracy and calibration for the classification tasks is investigated. The authors evaluate uncertainty not only for in-distribution examples but also for OOD examples. They utilize metrics such as negative log-likelihood and Brier scores to evaluate the model uncertainty or accuracy of computed predicted probabilities. Using large-scale experiments, the authors show that the calibration error increases with increasing distribution shift and post-hoc calibration does indeed fall short in detecting OOD examples.

\section{Intentional Anomaly Detection}

In this section, we discuss the detection techniques for detecting intentionally designed adversarial test examples given a pre-trained neural network. It is well known that DNNs are highly susceptible to test time adversarial examples -- human-imperceptible perturbations that, when added to any image, causes it to be misclassified with high probability~\cite{szegedy2013intriguing,goodfellow2014explaining}. 
The imperceptibility constraint ensures that the test example belongs to the data manifold yet gets misclassified. Hence, we need techniques to improve the reliability of the predictions or determine whether the test example is adversarial or normal. Here, we focus on the latter with the availability of a pre-trained DNN followed by a detector. Based on the availability of labels, the techniques are classified as supervised, semi-supervised, and unsupervised which are elaborated as follows and summarized in Table~\ref{tab:adversarialtable}.

\begin{table}[!t]
\caption{Adversarial example detection related papers.}
\begin{adjustbox}{width=\columnwidth,center}
 \begin{tabular}{||c c c||} 
 \hline
  Classification Type & Reference & Contributions  \\ [0.5ex] 
  \hline
  Supervised & \cite{metzen2017detecting} & \begin{tabular}{@{}c c@{}} Binary detector trained on\\ intermediate feature representations is \\proposed to detect adversarial examples\end{tabular}\\
      \hline
 Supervised  & \cite{Feinman2017DetectingAS} & \begin{tabular}{@{}c c@{}}Logistic regression based detector trained \\ with two features: the uncertainty\\  and the density estimate is used \end{tabular} \\ 
 \hline
Supervised & \cite{carrara2018adversarial} & \begin{tabular}{@{}c c@{}} LSTM based binary detector is trained\\ to analyze the sequence of deep features\\   embedded in a distance space \end{tabular}\\
\hline
Supervised & \cite{ma2018characterizing} &  \begin{tabular}{@{}c c@{}}  Local Intrinsic Dimensionality is used to \\characterize the dimensional properties of\\ the regions where the adversarial examples lie \end{tabular}\\
  \hline
  Supervised & \cite{Aigrain_corr_2019} & \begin{tabular}{@{}c c@{}} Three layer regression NN used as\\ the detector to predict confidence score\end{tabular}\\
  \hline
  Unsupervised  & \cite{Song_corr_2017} & \begin{tabular}{@{}c c@{}}Rank based statistics with generative models\\ is used for detecting adversarial examples \end{tabular}  \\
 \hline
Unsupervised & \cite{Miller_neuralcomputation_2017} & \begin{tabular}{@{}c c@{}} KL distance based metric  is applied \\on the posterior distributions to \\detect the adversarial examples\end{tabular}\\
\hline
Unsupervised & \cite{carrara2017detecting} & \begin{tabular}{@{}c c@{}} Nearest neighbor classification score\\ based on deep features is used as to\\ detect adversarial examples \end{tabular}\\
  \hline
     Unsupervised  & \cite{Zheng_nips_2018} & \begin{tabular}{@{}c c@{}}Adversarial examples are detected by \\ modeling output distribution of hidden\\ layers of the DNN given normal examples  \end{tabular}  \\ 
\hline
Unsupervised & \cite{ma2019nic}& \begin{tabular}{@{}c c@{}} Provenance and activation invariance \\ is used to detect adversarial examples \end{tabular}\\
\hline
Unsupervised & \cite{Sheikholeslami2019MinimumUB} & \begin{tabular}{@{}c c@{}} Mutual Information is used to detect\\ adversarial examples by minimizing\\ uncertainty over sampling probabilities \end{tabular}\\
\hline
 Unsupervised & \cite{dubey2019defense} & \begin{tabular}{@{}c c@{}} Detection by nearest neighbor search\\ based projections of adversarial examples\\ onto in-distribution image manifold is used  \end{tabular} \\ 
 \hline 
 Unsupervised & \cite{anirudh2019mimicgan} & \begin{tabular}{@{}c c@{}} Detection by gradient search based\\ projections of adversarial examples onto\\ in-distribution image manifold is used  \end{tabular}\\
\hline
\label{tab:adversarialtable}
\end{tabular}
\end{adjustbox}
\end{table}

\subsection{Supervised Approaches}

In this section, we discuss the detection techniques that require the labels of both in-distribution and adversarial examples and referred to them as supervised anomaly detection techniques. The test examples are compared against the detector to determine whether they are normal or adversarial.

In~\cite{metzen2017detecting}, a binary adversarial example detector is proposed. The detector is trained on intermediate feature representations of a pre-trained classifier on the original data set and adversarial examples. Although it may seem very difficult to train such a detector, their results on CIFAR10 and a 10-class subset of ImageNet datasets show that training such a detector is indeed possible. In fact, the detector achieves high accuracy in the detection of adversarial examples. Moreover, while the detector is trained on adversarial examples generated using a specific attack method, it is found that the detector generalizes to similar and weaker attack methods. Similar strategy was employed in~\cite{grosse2017statistical} where ML model was augmented with an additional class in which the model is trained to classify all adversarial inputs using labeled data.

The authors in~\cite{Feinman2017DetectingAS} proposed three methods to detect adversarial examples. First, method which is based on the density estimation uses estimates from the kernel density estimation of the training set in the feature space of the last hidden layer to detect adversarial examples. This method is meant to detect points that lie far from the data manifold. However, this strategy may not work well when adversarial example is very near the benign submanifold. Therefore, the authors proposed second approach which uses Bayesian uncertainty estimates from the dropout neural networks when points lie in low-confidence regions of the input space. They show that dropout based method can detect adversarial samples in situations where density estimates cannot. Finally, they also build a combined detector which is a simple logistic regression classifier with two features as input: the uncertainty and the density estimate. The combined detector is trained on a labeled training set which comprises of uncertainty values and density estimates for both benign and adversarial examples generated using different adversarial attack methods. The authors report that the performance of the combined detector (detection accuracy of 85-93\%) is better than detectors trained either on uncertainty or on density values, demonstrating that each feature is able to detect different qualities of adversarial features.

In~\cite{carrara2018adversarial}, the idea is that the trajectory of the internal representations in the forward pass for the adversarial examples are different from that of the in-distribution examples. The internal representations of an input is embedded into the feature distance spaces which capture the relative positions of an example with respect to a given in-distribution example in the feature space. The embedding enables compact encoding of the evolution of the activations through the forward pass of the network. Hence, facilitating the search for differences between the trajectories of in-distribution and adversarial inputs. An LSTM based binary detector is trained to analyze the sequence of deep features embedded in a distance space and detect adversarial examples. The experimental results show that the detection scheme is able to detect a variety of adversarial examples targeting the ResNet-50 classifier pre-trained on the ImageNet dataset.

In~\cite{ma2018characterizing}, an expansion-based measure of intrinsic dimensionality is used as an alternative to density measure to detect adversarial example. The expansion model of dimensionality assesses the local dimensional structure of the data and characterizes the intrinsic dimensionality as a property of the datasets. The Local Intrinsic Dimensionality (LID) generalizes this concept to the local distance distribution from a reference point to its neighbors -- the dimensionality of the local data submanifold in the vicinity of the reference point is revealed by the growth characteristics of the cumulative distribution function. The authors use LID to characterize the intrinsic dimensionality of regions where adversarial examples lie, and use estimates of LID to detect adversarial examples. Note that LID is a function of the nearest neighbor distances and it found to be significantly higher for the adversarial examples than the benign examples. A binary adversarial example detector is trained by using the training data to construct features for each sample, based on its LID across different layers, where the class label is assigned positive for adversarial examples and assigned negative for in-distribution examples. Experiments on several attack strategies show that LID based detector outperforms several state-of-the-art detection measures by large margins.

In~\cite{Aigrain_corr_2019}, a three layer regression NN is used as a detector that takes logits of in-distribution and adversarial examples from a pre-trained DNN as the input and predicts the confidence value, i.e., whether the classification is normal or adversarial. The classifier used is a pre-trained CNN trained using in-distribution datasets (MNIST and CIFAR) and the detector is trained on logits of both in-distribution and adversarial examples generated using different methods.  This work show that logits of a pre-trained network provide relevant information to detect adversarial examples.

\subsection{Semi-supervised Approaches}

Semi-supervised anomaly detection techniques utilize unlabeled contaminated data (or information) in addition to labeled instances of in-distribution class. Since, these techniques do not require to know whether unlabeled instance is in-distribution or adversarial examples, they are more widely applicable than supervised techniques. However, we could not find any existing semi-supervised adversarial example detection approach in the literature. Note that this may be a worthwhile direction to pursue in future research.

\subsection{Unsupervised Approaches}

We refer to the detection techniques as unsupervised if they only utilize in-distribution data for adversarial detection. 

In~\cite{Song_corr_2017}, the probabilities of all the training images under the generative model (such as, PixelCNN) is computed. Then, for a test example, the probability density at the input is computed and its rank among the density values of all the training examples is evaluated. This rank can be used as a test statistic which gives a $p$-value for whether the example is normal or adversarial. The method improves resilience of the state-of-the-art methods against attacks and increases the detection accuracy by a significant margin. Further, the authors suggest purifying adversarial examples by searching for more probable images within a small distance of the original training ones. By utilizing $L^{\infty}$ distance, the true labels of the purified images remains unchanged. The resulting purified images have higher probability under in-distribution so that the classifier trained on normal images will have more reliable predictions on these purified images. This intuition is used to build a more effective defense against adversarial attacks.

The motivation for the method in~\cite{Miller_neuralcomputation_2017} is that adversarial examples should be both (a) ``too atypical" (i.e.,  have atypically low likelihood) under the density model for the DNN-predicted class, and (b) ``too typical" (i.e., have too high a likelihood) under some class other than the DNN-predicted class. While it may seem that one requires to use two detection thresholds, they instead propose a single decision statistic that captures both requirements. Specifically, they define (a) a two-class posterior evaluated with respect to the (density-based) null model, and (b) corresponding two-class posterior evaluated via the DNN. Both deviations (``too atypical" and ``too typical") are captured by the Kullback-Leibler divergence decision statistic. A sample is declared adversarial if this statistic exceeds a preset threshold value.

The approach in~\cite{carrara2017detecting} performs a kNN similarity search among the deep features obtained from the training images to a given test image classified by the DNN. They then use the score assigned by a kNN classifier to the class predicted by the DNN as a measure of confidence of the classification. Note that this approach does not rely on the classification produced by the kNN classifier, but only use the score assigned to the DNN prediction as a measure of confidence. The intuition behind this approach is that while it is unlikely that a class correctly predicted by the DNN has the highest kNN score among the scores of all the classes, it is implausible that a correct classification has a very low score. Results on the ImageNet dataset show that hidden layers activations can be used to detect misclassifications caused by various attacks.

In~\cite{Zheng_nips_2018}, intrinsic properties of the pre-trained DNN, i.e., output distributions of the hidden neurons, are used to detect adversarial examples. Their motivation is that when the DNN incorrectly assigns an adversarial example to a specific class label, the distribution of its hidden states are very different as compared to those obtained by the normal data of the same class. They use Gaussian Mixture Model (GMM) to approximate the hidden state distribution of each class using benign training data. Likelihoods are then compared to the respective class thresholds to detect whether an example is adversarial or not. Experimental results  on standard datasets (MNIST, F-MNIST, CIFAR-10) against several attack
methods show that this approach can achieve state-of-the-art robustness in defending black-box and gray-box attacks.

The authors in~\cite{ma2019nic} found that adversarial examples  mainly exploit two attack channels: the provenance channel and the activation value distribution channel. The provenance channel imply instability of DNN output to small changes in activation values, which eventually leads to misclassification. On the other hand, the activation channel imply that while the provenance changes slightly, the activation values of a layer may be substantially different from those in the presence of benign inputs. Exploiting these observations they propose a method that extracts two kinds of invariants (or probability distributions denoted by
models), the value invariants to guard the value channel and the provenance invariants to guard the provenance channel. This is achieved by training a set of models for individual layers to describe the activation and provenance distributions only using in-distribution inputs. In other words, invariant models are trained as a One-Class Classification (OCC) problem where all training samples are positive (i.e., in-distribution inputs in this context). At test time, an input is passed through all the invariant models which provide independent predictions about whether the input induces states that violate the invariant distributions. The final result is a joint decision based on all these predictions. Extensive experiments on various attacks, datasets and models suggest that this method can achieve consistently high detection accuracy on all different types of attacks, while the performance of baseline detectors is not consistent.

In~\cite{Sheikholeslami2019MinimumUB}, the idea is that inherent distance of adversarial perturbation from the training data manifold will cause the overall network uncertainty to exceed that of the normal example. To this end, random sampling of hidden units of each layer of a pre-trained network is used to introduce randomness and the overall uncertainty of a test image is quantified in terms of the hidden layer components. A mutual information based thresholding test is used to detect adversarial examples. The performance is further improved by optimizing over the sampling probabilities to minimize uncertainty. Experiments on the CIFAR10 and the cats-and-dogs datasets on deep state-of-the-art CNNs demonstrated the importance sampling parameter optimization, which readily translate to improved attack detection.

Approaches such as \cite{dubey2019defense} and \cite{anirudh2019mimicgan} rely on projecting the test image to benign dataset manifold to detect adversarial examples. The underlying assumption in these approaches is that adversarial perturbations move the test image away from the benign image manifold and the effect of adversary can be nullified by projecting the images back onto the benign manifold before classifying them. As the true image manifold is unknown, various estimation techniques are used. For example, \cite{dubey2019defense} use a sample approximation comprising a database of billions of natural images. On the other hand,  \cite{anirudh2019mimicgan} use a generative model trained on benign images to estimate the manifold. 
Given the estimated benign manifold, the projection is done by nearest neighbor search in \cite{dubey2019defense} and gradient-based search in \cite{anirudh2019mimicgan}. These methods are founds to be robust against gray-box and black-box attacks where the adversary is unaware of the defense strategy. 


\subsection{Other Miscellaneous Techniques}

Here we discuss some other techniques that are used for adversarial example detection which do not fall in the aforementioned categorizations of the post-hoc processing.

In~\cite{smith2018understanding}, various uncertainty measures, e.g.,  entropy, mutual information, softmax variance, for adversarial example detection are examined. Each of these measures capture distinct types of uncertainty and are analyzed from the perspective of adversarial example detection. The authors showed that only the mutual information gets useful detection performance on adversarial examples. In fact, most other measures of uncertainty seem to be worse than random guessing on MNIST and Kaggle dogs vs. cats classification datasets.

The approach in~\cite{Xu_corr_2017} is motivated by the observation that the DNN feature spaces are often unnecessarily large, and this provides extensive degrees of freedom for an attacker to construct adversarial examples. The authors propose to reduce the degrees of freedom for constructing adversarial examples by ``squeezing" out unnecessary input features. Specifically, they compare the  model's prediction of the original test example with its prediction of the test example after squeezing, i.e., reducing the color depth of images, and using smoothing to reduce the variation among pixels. If the original and the squeezed inputs produce substantially different predictions then the example is declared adversarial.

In~\cite{Lu2017SafetyNetDA} SafetyNet is proposed which consists of the original classifier, and an adversary detector which looks at the internal state of the later layers in the original classifier. Here, the output from the ReLU is quantized to generate a discrete code based on some set of thresholds. They claimed that different code patterns appear for natural examples and adversarial examples. An adversarial example detector (i.e., RBF-SVM) is used that compares a code produced at test time with a collection of examples, i.e., an attacker must make the network produce a code that is acceptable to the detector which is shown to be hard.

In~\cite{li2018generative}, the method improves the naive Bayes used in many generative classifiers by combining it with variational auto-encoder. They propose three adversarial example detection methods. The first two use the learned generative model as a proxy of the data manifold, and reject inputs that are far away from it. The third computes statistics for the classifier’s output probability vector, and rejects inputs that lead to under-confident predictions. Experimental results suggest that deep Bayes classifiers are more robust than deep discriminative classifiers, and that the detection methods based on deep Bayes are effective against various attacks.

In~\cite{ahuja2019probabilistic}, the authors propose to model the outputs of the various layers (deep features) with parametric probability distributions (Gaussian and Gaussian Mixture Models).  At test time, the log-likelihood scores of the features of a test sample are calculated with respect to these distributions and used as anomaly score to discriminate in-distribution samples (which should have high likelihood) from adversarial examples (which should have low likelihood).

The main idea in~\cite{cohen2019detecting} is to combine kNN based distance measure~\cite{sitawarin2019defending} with influence function which is a measure of how much a test sample classification is affected by each training sample. The motivation behind this approach is that for an in-distribution input, its kNN training samples (nearest neighbors in the embedding space) and the most helpful training samples (found using the influence function) should correlate. However, this correlation is much weaker for adversarial examples, and serves as an indication of the attack.

The motivation in~\cite{monteiro2019generalizable} is the observation that different neural networks presented with the same adversarial example will make different mistakes. The authors propose to use such mistake patterns for adversarial example detection. Experiments on the  MNIST and CIFAR10 datasets show that such detection approach generalizes well across different adversarial example generation methods.

In~\cite{freitas2020unmask} robust feature alignment is used to detect adversarial examples. By using an object detector, the authors first extract higher-level robust features contained in images. Next, the approach quantifies the similarity between the image's extracted features with the expected features of its predicted class. A similarity threshold is finally used to classify a test sample as benign or adversarial.

In~\cite{Wang_corr_2016}, anomaly detection is performed by introducing random feature nullification in both training and testing phases that ensures the non-deterministic nature of the DNN. Here, the randomization introduced at the test time ensures that the model’s processing of the input decreases the effectiveness of the adversarial examples even if the attacker learns critical features.

In~\cite{Pertigkiozoglou_corr_2018}, three strategies are presented. First, regularized feature vectors are used to retrain the last layer of the CNN. This can be used to detect whether the input is adversarial. Second,  histograms are created from the absolute values of the hidden layer outputs and are combined to form a vector which is used by the SVM to classify. Third, the input is perturbed to reinforce the parts of the input example that are ignored by the DNN which can then be used for adversarial example detection. Finally, the authors combine the best aspects of these methods to develop a more robust approach.

In~\cite{Li_corr_2018}, a framework is presented for enhancing the robustness of DNN against adversarial examples. The idea is to use locality-preserving hash functions to transform examples to enhance the robustness. The hash representations of the examples are reconstructed by using a denoising auto-encoder (DAE) that enables the DNN classifier to attain the locality information in the latent space. Moreover, the DAE can detect the adversarial examples that are far from the support of the underlying training distribution.

\section{Relative Strengths and Weakness}
\label{pros}

The supervised techniques usually have higher performance compared to other methods as they use the labeled examples from both normal and anomaly classes. They are able to learn the boundary from the labeled training examples and then more easily classify the unseen test examples into normal or anomaly classes. However, when training data for anomalies (the known unknowns) may not represent the full spectrum of anomalies, supervised approaches may overfit and perform poorly on unseen anomalous data (the unknown unknowns). Furthermore, due to the lack of availability of labeled anomalous examples, supervised techniques are not as popular as the semi-supervised or unsupervised techniques. 

Unsupervised techniques are quite flexible and broadly applicable as they do not rely on the availability of the anomalous data and corresponding labels. The techniques learn inherent characteristics or unique features solely from in-distribution data that are useful in separating normal from anomalous examples. Unfortunately, this flexibility comes at the cost of robustness -- the unsupervised techniques are very sensitive to noise, and data corruptions and are often less accurate than supervised or semi-supervised techniques.

Semi-supervised techniques exploit unlabeled data in addition to labeled in-distribution data to improve the performance of unsupervised techniques. Though, whether unlabeled data is in-distribution or anomaly is not known, it is observed that unlabeled data is helpful in improving the performance of anomaly detection. Note that unlabeled data can be obtained easily in real-world applications making semi-supervised techniques amenable in practice. These methods also suffer from the overfitting problem on unseen anomalies.  

Distance-based methods, e.g., kNN approaches, require appropriate distance measure to be defined a priori. Most distance measures are not effective in high-dimension. Further, such methods are typically heuristic and require manual selection of parameters. Projection-based methods, e.g., GAN approaches, are very flexible and address the high-dimensionality challenge. However, their performance is heavily dependent on the quality of the image manifold estimate. In certain applications, it may not be easy to estimate the image manifold with sample approximation or generative modeling. Probabilistic methods, e.g., density estimation approaches, make use of the distribution of the training data or features to determine the location of the anomaly boundary. The performance of such methods is very poor in the small data regime as reliable estimates cannot be obtained. Uncertainty-based methods, e.g., entropy approaches, require a metric that is sensitive enough to detect the effects of anomalies in the dataset. Although these methods are easy to implement in practice, the performance of such methods is highly dependent on the the quality of uncertainties. Uncertainty quantification in DL is an ongoing research topic and high quality uncertainty estimates will surely improve the performance of uncertainty-based methods.

The computational complexity of these methods is another important aspect to consider. In general, probabilistic and uncertainty-based methods have computationally expensive training phases, however  efficient testing. On the other hand, distance-based and projection-based methods, in general, are computationally expensive in the test phase.
Depending on the application requirements, a user should choose the most appropriate anomaly detection method.

\section{Application Domains}
In this section, we briefly discuss several applications of OOD and adversarial example detection. We also suggest future research that is needed for these application domains.

\vspace{0.05in}
{\bf{Intrusion Detection}} - An Intrusion Detection System is a system that monitors network traffic for suspicious activity and issues alerts when such activity is discovered. A key challenge for intrusion detection is the huge volume of data and sophisticated malicious patterns. Therefore, DL techniques are quite
promising in the intrusion detection application.

In~\cite{Ryan_nips_1998}, a neural network based intrusion detector is trained to identify intruders. In~\cite{Gao_cacb_2014}, a deep hierarchical model is proposed for intrusion detection. The model is a combination of a restricted Boltzmann machine (RBM) for unsupervised feature learning and a supervised learning network called as Backpropagation network.
In~\cite{Yu2017NetworkID}, a network intrusion model is proposed where feature learning is performed by stacking dilated convolutional autoencoders. These feature are then used to train a softmax classifier to perform supervised intrusion detection. 
In~\cite{Aygun_csc_2017}, an autoencoder based model in combination with a stochastic anomaly threshold determination method is proposed for intrusion detection. The algorithm computes the threshold using the empirical mean and standard deviation which are found from training set via the trained autoencoder.

As mentioned earlier, these DL based systems are equally susceptible to both OOD and adversarial examples~\cite{Wang_access_2018, int1, int2}. In~\cite{Wang_access_2018}, the authors analyze the performances of the state-of-the-art attack algorithms against DL-based intrusion detection. The susceptibility of DNNs used in the intrusion detection system is validated by experiments and the role of individual features is also explored. The authors in~\cite{int1} demonstrated that an adversary can generate effective adversarial examples against DL based intrusion detection systems even when the internal information of the target model is not available to the adversary. Note that in intrusion detection applications, a large amount of labeled data corresponding to normal behavior is usually available, while labels for intrusions are not. Therefore, semi-supervised and unsupervised OOD and adversarial example detection techniques discussed in the previous sections are worthwhile directions to pursue. 

\vspace{0.05in}
{\bf Fraud Detection} - Fraud detection refers to detection of fraudulent activities occurring in many e-commerce domains, such as, banking, insurance, law enforcement, etc.
A good fraud detection system should be able to identify the fraudulent transactions accurately and should make the detection possible in real-time. There is an increase in interest in applying DL techniques in fraud detection systems. 
In~\cite{Jurgovsky2018SequenceCF}, fraud detection is modeled as a sequence classification task. An LSTM is used to generate transaction sequences and incorporate aggregation functions like mean, absolute value to aggregate the learned features for fraud detection. Furthermore, in~\cite{Zhang2018AMB}, feature sequencing is performed using CNNs for detecting transaction fraud.  
Recently, the authors in~\cite{guo2019securing} analyzed the
vulnerability of deep fraud detector to adversarial examples, i.e., slight perturbations in input transactions designed to fool the fraud detector. They show that the deployed deep fraud detector is highly vulnerable to attacks as the average precision is decreased from 90\% to as low as 20\%.

This motivates the study of the effect of unintentional and intentional anomalies in deep fraud detection systems. Techniques discussed in the previous sections will be applicable for such a problem and are potential viable solutions for designing robust deep fraud detection systems.     

\vspace{0.05in}
{\bf Anomaly Detection in Healthcare and Industrial Domains} - Anomaly detection in the healthcare domain try to detect abnormal patient conditions or instrumentation errors. Anomaly detection is a very critical problem in this domain and requires high degree of accuracy. Similarly, in industrial systems like wind turbines, power plants, and storage devices which are exposed to large amounts of stress on a daily basis, it is critical to detect any damages as quickly as possible. The medical abnormalities and industrial damage are rare events and detecting them can be modeled as an anomaly detection problem. Therefore, there is a surge of interest in applying DL in both medical~\cite{esteva2019guide} and industrial application domains~\cite{nash2018review}. 

Unfortunately, similar to other DL applications, these systems are equally susceptible to OOD and adversarial examples. For example, the authors in~\cite{finlayson2018adversarial}  
demonstrated that adversarial examples are capable of manipulating DL systems across three clinical domains: diabetic retinopathy from retinal fundoscopy, pneumothorax from chest-Xray, and melanoma from dermoscopic photographs. 

This motivates the study of the effect of anomalies in DL based healthcare and industrial systems. Techniques discussed in the previous sections can be used for designing  robust healthcare and damage detection systems.

\vspace{0.05in}
{\bf Malware Detection} - Malware detection focuses on detecting malware software by monitoring the activity of the computer systems and classifying it as normal or anomalous. The velocity, volume, and the complexity of malware are posing new challenges to the anti-malware community. Current state-of-the-art research shows that recently, researchers started applying machine learning and DL methods for malware analysis and detection~\cite{rathore2018malware}. In~\cite{Wang_corr_2016}, malware detection is performed by introducing random feature nullification in both training and testing phases that ensures the non-deterministic nature of the DNNs. Intuitively, the non-deterministic nature ensures that the model’s processing of the input decreases the effectiveness of the adversarial examples even if the attacker learns critical features. Furthermore, in~\cite{Hardy2016DL4M}, a stacked autoencoders  model is used for malware detection. The model employs a greedy layerwise training operation for unsupervised feature learning and supervised parameter tuning. Furthermore, in~\cite{Kim2018ZerodayMD}, fake malware is generated and is learned to distinguish from the real data using a novel GAN architecture. 

Authors in \cite{grosse2017adversarial,suciu2019exploring} expanded on existing adversarial example crafting algorithms to construct a highly-effective attack against malware detection models. Using the augmented adversarial crafting algorithm, authors managed to mislead the malware detection classifier for 63\% of all malware samples. In~\cite{Li_corr_2018}, the authors analyzed the effect of several attacks on the Android malware classification task.

Given the susceptibility of the state-of-the-art malware detection classifiers to adversarial examples, it will be useful to utilize OOD and adversarial example detection techniques in deep malware detection systems.

\vspace{0.05in}
{\bf Time Series and Video Surveillance Anomaly Detection} - The task of detecting anomalies in multivariate time series data is quite challenging. Hence, efficient detection of multivariate time series anomalies is critical for fault diagnostics. RNN and LSTM based methods perform well in detecting anomalies in multivariate time series data. In~\cite{Buda_kddm_2018}, a generic framework based on DL for detecting anomalies in multivariate time series data is presented. Deep attention based models are used in~\cite{Yuan_icdm_2018} for anomaly detection for effective detection of anomalies. Many works have applied the deep learning models for video surveillance anomaly detection in~\cite{Gutoski2017DetectionOV,Benari_corr_2018,Tripathi2019}.

Unfortunately, some recent papers~\cite{ts1,ts2} have shown that one can design adversarial examples on time-series classifiers as well. Thus, in our opinion, future researchers should incorporate OOD and adversarial example detectors in their time series classification systems to improve the resilience and consider model robustness as an evaluative metric.

\section{Conclusion and Open Questions}
In this survey, we discussed various techniques for detecting OOD and adversarial examples given a pre-trained DNN. 
For each category of anomaly detection techniques, we discussed the strengths and weaknesses of these techniques.
Finally, we discussed various application domains where the post-hoc processing, as well as, training based anomaly detection techniques are applicable. 

There are several open issues and worthwhile future directions for further research. Several of these are identified by analyzing and comparing existing literature and the research considered in this survey.

\noindent \textbf{Methods:} We classified anomaly detection algorithms based on the availability of the labels of anomalous examples and the type of metrics used. Based on the availability of the labels, the techniques are classified as supervised, semi-supervised, and unsupervised.
Based on the type of metric, the techniques are classified as probability-based, distance-based, projection-based, and uncertainty-based.
Each category of methods have their own strengths and weaknesses, and faces different challenges as discussed in Section~\ref{pros}. We conjecture that exploration of ensemble detection approaches can be a worthwhile future direction. 
The ensemble approach combines outputs of multiple detectors offering complementary strengths into a single one, thus yielding better performance compared to using individual detectors. 

\noindent \textbf{Defining Anomalies:} Majority of the research on detecting OOD and adversarial examples in DL focuses on detecting independent anomalies (e.g., adversarial examples generated independently from one another). However, anomalous behaviors can be much more complex requiring more sophisticated detection approaches than currently available. An example of this is discussed in \cite{goodfellow2019research} where a simple correlated anomaly generation approach was discussed. It was shown that current defenses are not capable of defending against this simple scheme. Further, defining collective and contextual anomalies~\cite{zhang2018spectral} in the context of OOD and adversarial examples in DL can be very interesting and detecting them will certainly require the development of a new class of detectors. Also, we want to emphasize that it is important for future research on anomaly detection to be cognizant of the fact that anomalies may not adhere to our definitions and assumptions and can have extremely complex unknown behavior. This is similar to the concept of unknown-unknowns~\cite{lakkaraju2017identifying}. We believe that the research on domain generalization~\cite{muandet2013domain} and meta learning~\cite{vilalta2002perspective} can be used to solve some of these issues. 

\noindent\textbf{Going beyond image classification:} Most of the papers discussed in this survey (and in the literature) focus on the detection of anomalous examples in DNN based image classification problems. However, in recent years there has been a surge of interest in applying DL on other data types, e.g. text, graphs, trees, manifolds etc. These data types are ubiquitous in several high-impact applications including bioinformatics, neuroscience, social sciences, and molecular chemistry.  Unfortunately, DL approaches in these data types also suffer from the existence of OOD and adversarial examples~\cite{xu2019adversarial}. Post-hoc detection of such anomalies has not received much attention. Furthermore, going beyond classification problems and exploring the design and the detection of anomalies in DL based object detection, control, and planning problems can be a high-impact future research direction.  

\noindent \textbf{Performance Evaluation:} Reliably evaluating the performance of OOD and adversarial example detection methods has proven to be extremely difficult. Previous evaluation methods are found to be ineffective and performing incorrect or incomplete evaluations~\cite{carlini2017adversarial, shafaei2018less}. Absence of a standard definition for anomalies makes this problem very challenging. Furthermore, as anomalies become more sophisticated, it may become even harder to reliably evaluate the detection performance. Majority of current approaches evaluate the performance of anomaly detectors on OOD and adversarial examples. Assuming that training data may not represent the full spectrum of anomalies, this evaluation approach raises the risk of overfitting. Ideally, one should adopt an evaluation method that can assess the detection performance on adaptive and unseen anomalies (the unknown unknowns) over methods that only can assess the detection performance on previously seen anomalies (the known unknowns). Due to these reasons, there is an immediate need for designing principled benchmarks to reliably evaluate the anomaly detection performance~\cite{shafaei2018less,carlini2019evaluating}.

\noindent\textbf{Theoretical analysis and Fundamental Limits:} Finally, we need to make efforts on the theoretical front to understand the nature of the anomaly detection problem in DL-based systems. In the recent past, a pattern has emerged in which the majority of heuristics based defenses (both posthoc detection and training based) are easily broken by new attacks~\cite{athalye2018obfuscated, carlini2017adversarial}.
Therefore, the development of a coherent theory and methodology that guides practical design for anomaly detection in DL-based systems~\cite{vernekar2019analysis}, and fundamental characterizations of the existence of adversarial examples~\cite{shafahi2018adversarial} is of utmost importance.  
How to leverage special learning properties such as the spatial and temporal consistencies to identify OOD examples~\cite{xiao2018characterizing,yang2018characterizing} also worth further exploration.

To summarize, OOD and adversarial example detection in DL-based systems is an open problem. We highlighted several aspects of the problem to be understood on both theoretical and algorithmic front to improve the effectiveness and feasibility of anomaly detection. We hope that this survey will provide a comprehensive understanding of the different approaches, show the bigger picture of the problem, and suggest few promising directions for researchers to pursue in further investigations on the anomaly detection in DL-based systems.

\section*{Acknowledgement}
This work was performed under the auspices of the U.S. Department of Energy by Lawrence Livermore National Laboratory under Contract DE-AC52-07NA27344.

{ \bibliographystyle{IEEEtran}
  \bibliography{Refs}}

\end{document}